%% file: main.tex
\documentclass[runningheads]{llncs}
\pdfoutput=1    

\usepackage{eccv}

\usepackage{svg}
\usepackage{eccvabbrv}

\usepackage{pifont}
\newcommand{\cmark}{\ding{51}}%
\newcommand{\xmark}{\ding{55}}%

\usepackage{subcaption}
\usepackage{graphicx}
\usepackage{booktabs}
\graphicspath{{img/}}

\usepackage[accsupp]{axessibility}  

\usepackage{hyperref}

\usepackage{orcidlink}

\begin{document}
\title{AnoVox: A Benchmark for Multimodal\\ Anomaly Detection in Autonomous Driving} 

\titlerunning{AnoVox: A Benchmark for Anomaly Detection}

\author{Daniel Bogdoll\inst{1,2}*\orcidlink{0000-0003-0432-4937} \and
Iramm Hamdard\inst{1,2}* \and
Lukas Namgyu Rößler\inst{2}* \and
Felix Geisler\inst{2} \and
Muhammed Bayram\inst{2} \and
Felix Wang\inst{2} \and
Jan Imhof\inst{1,2} \and
Miguel de Campos\inst{3} \and
Anushervon Tabarov\inst{1,2} \and
Yitian Yang\inst{1} \and
Martin Gontscharow\inst{1} \and 
Hanno Gottschalk\inst{3}\orcidlink{0000-0003-2167-2028} \and
J. Marius Zöllner\inst{1,2}\orcidlink{0000-0001-6190-7202}
}

\authorrunning{Bogdoll et al.}

\institute{FZI Research Center for Information Technology, 76131 Karlsruhe, Germany
\email{bogdoll@fzi.de}\\
\and
Karlsruhe Institute of Technology, 76131 Karlsruhe, Germany
\and
Technical University of Berlin, 10623 Berlin, Germany
}

\maketitle

\def\thefootnote{\textsuperscript{\textasteriskcentered}}\footnotetext{These authors contributed equally}\def\thefootnote{\arabic{footnote}}

\begin{abstract}
\input{sec/0_abstract}
\end{abstract}
\input{sec/1_intro}
\input{sec/2_sota}

\input{sec/3_data}
\input{sec/4_eval}
\input{sec/5_conclusion}

\input{sec/6_acknowledgment}

\bibliographystyle{splncs04}
\bibliography{egbib}
\end{document}

%% file: sec/0_abstract.tex
The scale-up of autonomous vehicles depends heavily on their ability to deal with rare objects on the road. In order to handle such situations, it is necessary to detect anomalies in the first place. Anomaly detection has made great progress in the past years but suffers from poorly designed benchmarks with a strong focus on camera data. In this work, we present AnoVox, the largest benchmark for ANOmaly detection in autonomous driving to date. AnoVox incorporates multimodal sensor data and spatial VOXel ground truth, allowing for the comparison of methods independent of their used sensor. We propose a formal definition of normality and provide a compliant training dataset. AnoVox is the first benchmark to contain both \textit{content} and \textit{temporal} anomalies.

%% file: sec/1_intro.tex
\section{Introduction}
\label{sec:introduction}

In the past years, we have seen autonomous vehicles scale from small Operational Design Domains (ODD) to city-wide areas, reaching geographic areas covering multiple neighboring cities in the near future~\cite{waymo_one_2018, waymo_one_2024, waymo_peninsula}. However, with a growing fleet in an expanding ODD, scenarios from the long tail of rare events~\cite{Urtasun_cvpr_wad} occur more frequently as a consequence of increased exposure. Expert perspectives, focusing on content and temporal anomalies, are commonly used to judge "long tail" data points~\cite{Breitenstein_Systematization_2020_IV, breitensteinCornerCasesVisual2021, Heidecker_Application_2021_IV, Bogdoll_Description_2022_ICCV, Bogdoll_Ontology_2022_ECCV,pfeilWhySystemMakes2022,Roesch_Space_2022_SSCI}, as replaying training data during inference is not possible. Such expert views are often used for the creation of anomaly detection benchmarks but are of no help for anomaly detection. Few publications take a data-oriented perspective on the definition of anomalies~\cite{heideckerCornerCasesMachine2024, Chan_Detecting_2022_Springer,zhou_corner_data}, also known as out-of-distribution (OOD), corner cases, or outliers. Recent research has focused on anomaly detection methods \cite{hendrycks2016baseline} and its ramifications, such as OOD segmentation~\cite{NUNES2023296}, object detection~\cite{Du_Unknown_2022_CVPR}, instance segmentation~\cite{mohan2023panoptic}, and video tracking~\cite{Maag_Video_2022_ACCV}. These methods typically do not learn a representation of normality from training data alone, but require a semantic segmentation stage and fine-tuning on auxiliary data containing (artificial) anomalies~\cite{Blum_Fishyscapes_2021_IJCV,Chan_SegmentMeIfYouCan_2021_NEURIPS}. While there exists a considerable number of datasets in autonomous driving~\cite{liu2024survey,li2024opensourced,Bogdoll_Addatasets_2022_VEHITS}, only around 5 - 10\% of datasets are designed for anomaly detection~\cite{bogdoll_perception,Bogdoll_Anomaly_2022_CVPR}. These existing datasets, however, have significant limitations. First, the majority are camera-only, despite autonomous driving relying on multi-modal sensor setups. Second, temporal information is often overlooked, limiting anomaly detection to single frames. Additionally, \textit{content} anomalies are rarely found in challenging traffic situations, which allows for the assumption that anything on the road might be an anomaly~\cite{Maag_Video_2022_ACCV}. Most anomalies are human-defined, such as dogs on the street~\cite{Blum_Fishyscapes_2021_IJCV}, making detection challenging as these classes are typically included in training datasets but not annotated~\cite{cordts_cityscapes_2016,Du_Unknown_2022_CVPR}. As a result, anomaly detection methods can miss anomalies they don't perceive as atypical due to their training.


\begin{figure}[tbp]
    \centering
    \includegraphics[width=\textwidth]{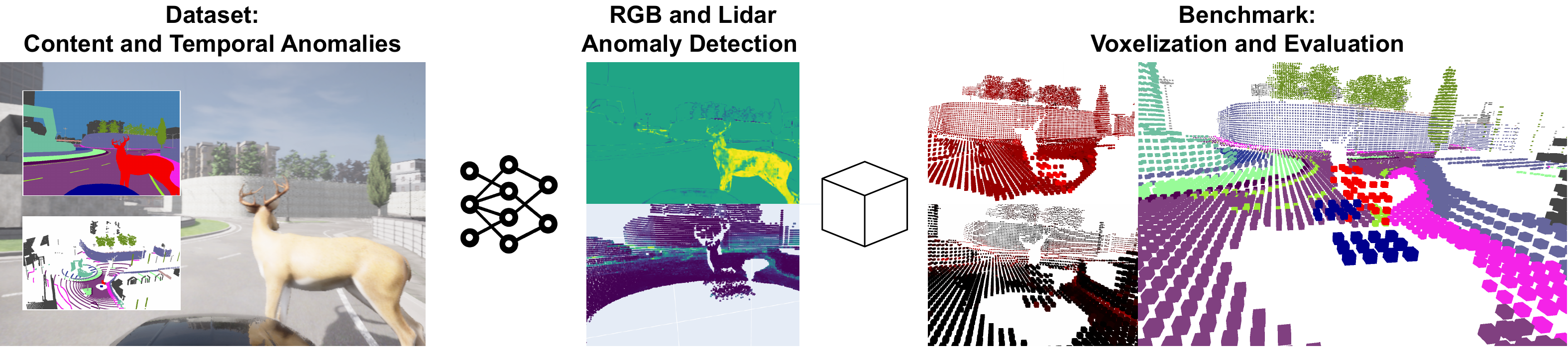}
    \caption{\textbf{Overview of AnoVox.} AnoVox includes both \textit{content} and \textit{temporal} anomalies. Here, we see an example of a \textit{content} anomaly. Ground truth is available for both camera and lidar, enabling the application of anomaly detection methods for both modalities. We additionally provide ground truth in a voxelized form. By mapping the anomaly scores to this voxelized space, we can compare methods from any sensor.}
    \label{fig:teaser}
\end{figure}

With AnoVox, we offer a challenging and adaptable benchmark for ANOmaly detection, leveraging the CARLA simulation engine~\cite{Dosovitskiy2017CARLAAO} to provide ground truth in all modalities and a spatial VOXel representation. Given the surge in foundation models~\cite{caron2021emergingdino,oquab2023dinov2}, world models~\cite{zhangLearningUnsupervisedWorld2023,huGAIA1GenerativeWorld2023}, and large-scale pre-training approaches~\cite{Hess_2023,yang2023unipad}, we see potential for the application of self- and unsupervised training paradigms in the field of anomaly detection as well. Hence, we prioritize support for label-free training paradigms and provide an extensive synthetic dataset to enable research in OOD segmentation, addressing the limitations of current datasets. All code is available on \href{https://github.com/fzi-forschungszentrum-informatik/anovox}{GitHub}. The dataset can be found on \href{https://zenodo.org/communities/anovox/}{Zenodo}. Our contributions include:

\begin{enumerate}
    \item Formal definition of normality and provision of a compliant training dataset for a fair comparison of anomaly detection methods
    \item Large and challenging benchmark with a high scene complexity and temporal scenarios that include many frames without anomalies
    \item First benchmark to provide content and temporal anomalies, adaptable multimodal sensor data, ego-vehicle state information, and ground truth in all modalities, as well as a spatial voxel representation
\end{enumerate}

%% file: sec/2_sota.tex
\section{Related Work}
\label{sec:sota}

\begin{table*}[t!]
    \caption{Vision-based anomaly detection benchmarks. In the \textit{Normality} column, $^\dagger$ denotes a domain shift between normal data and the proposed dataset, and $^\circledast$ denotes that the data with anomalies is based on a subset of the normal data. Adapted from~\cite{bogdoll_perception}.}
    \resizebox{\textwidth}{!}{%
        \begin{tabular}{@{}llccccccccr@{}}
            \toprule
            \textbf{Dataset}                                                                  & \textbf{Data}              & \textbf{Size} & \begin{tabular}[c]{@{}c@{}}\textbf{Ano.}\\\textbf{Source}\end{tabular} & \begin{tabular}[c]{@{}c@{}}\textbf{Ano.}\\\textbf{Type}\end{tabular} & \begin{tabular}[c]{@{}c@{}}\textbf{\#Ano.}\\\textbf{Classes}\end{tabular} & \begin{tabular}[c]{@{}c@{}}\textbf{Ground}\\\textbf{truth}\end{tabular} & \begin{tabular}[c]{@{}c@{}}\textbf{Temp.}\\\textbf{Data}\end{tabular} & \begin{tabular}[c]{@{}c@{}}\textbf{Ego}\\\textbf{Act.}\end{tabular} & \begin{tabular}[c]{@{}c@{}}\textbf{Reg.}\\\textbf{Tasks}\end{tabular} & \textbf{Normality}             \\ \midrule
            \textbf{Fishyscapes}~\cite{Blum_Fishyscapes_2019_ICCV,Blum_Fishyscapes_2021_IJCV} &                               &                                                                            &                                                                        &                                                                      &                                                                           &                                                                         &                                                                       &                                                                     &                                                                                                        \\
            FS Lost and Found                                                                 & Cam.                        & 375                                                                  & Recording                                                              & Content                                                              & 1                                                                         & Sem. mask (2D)                                                           & \textbf{---}                                                          & \textbf{---}                                                        & \textbf{---}                                                          & \textbf{---}                   \\
            FS Static                                                                         & Cam.                        & 1,030                                                                 & Data Augmentation                                                      & Content                                                              & 1                                                                         & Sem. mask (2D)                                                           & \textbf{---}                                                          & \textbf{---}                                                        & \textbf{---}                                                          & Cityscapes$^\circledast$       \\ \midrule
            \textbf{Crash to Not Crash}~\cite{kimCrashNotCrash2019}                           &                               &                                                                            &                                                                        &                                                                      &                                                                           &                                                                         &                                                                       &                                                                     &                                                                                                        \\
            YouTubeCrash                                                                      & Cam.                        & 2,400                                                                      & Web Sourcing                                                           & Temporal                                                             & 1                                                                         & Bbox (2D)                                                     & \cmark                                                                & \textbf{---}                                                        & \cmark                                                                & YouTubeCrash                   \\
            GTACrash                                                                          & Cam.                        & 154,400                                                                    & Simulation                                                             & Temporal                                                             & 1                                                                         & Bbox (2D)                                                     & \cmark                                                                & \textbf{---}                                                        & \cmark                                                                & \textbf{GTA V}                 \\ \midrule
            \textbf{CAOS}~\cite{Hendrycks_Scaling_2022_ICML}                                  &                               &                                                                            &                                                                        &                                                                      &                                                                           &                                                                         &                                                                       &                                                                     &                                                                       &                                \\
            StreetHazards                                                                     & Cam.                        & 1,500                                                                      & Simulation                                                             & Content                                                              & 1                                                                         & Sem. mask (2D)                                                           & \cmark                                                                & \textbf{---}                                                        & \cmark                                                                & CARLA$^\dagger$                \\
            BDD-Anomaly                                                                       & Cam.                        & 810                                                                        & Class Exclusion                                                        & Content                                                              & 3                                                                         & Sem. mask (2D)                                                           & \textbf{---}                                                          & \textbf{---}                                                        & \cmark                                                                & BDD100K$^\circledast$          \\ \midrule
            \textbf{SegmentMeIfYouCan}~\cite{Chan_SegmentMeIfYouCan_2021_NEURIPS}             &                               &                                                                            &                                                                        &                                                                      &                                                                           &                                                                         &                                                                       &                                                                     &                                                                       &                                \\
            RoadAnomaly21                                                                     & Cam.                        & 110                                                                   & Web Sourcing                                                           & Content                                                              & 1                                                                         & Sem. mask (2D)                                                           & \textbf{---}                                                          & \textbf{---}                                                        & \textbf{---}                                                          & Cityscapes$^\dagger$           \\
            RoadObstacle21                                                                    & Cam.                        & 412                     & Recording                                                              & Content                                                              & 1                                                                         & Sem. mask (2D)                                                          & \cmark                                                                & \textbf{---}                                                        & \textbf{---}                                                          & Cityscapes$^\dagger$           \\ \midrule
            \textbf{Rare Road Objects}~\cite{buCARLASimulatedData2021}                        &                               &                                                                            &                                                                        &                                                                      &                                                                           &                                                                         &                                                                       &                                                                     &                                                                       &                                \\
            Synthetic Fire Hydrants
                                                                                              & Cam.                        & 30,000                                                                     & Simulation                                                             & Content                                                              & 1                                                                         & Bbox (2D)                                                     & \textbf{---}                                                          & \textbf{---}                                                        & \textbf{---}                                                          & CARLA$^\dagger$                \\
            Synthetic Crosswalks
                                                                                              & Cam.                        & 20,000                                                                     & Simulation                                                             & Content                                                              & 1                                                                         & Bbox (2D)                                                     & \textbf{---}                                                          & \textbf{---}                                                        & \textbf{---}                                                          & CARLA$^\dagger$                \\ \midrule
            \textbf{CODA}~\cite{Li_CODA_2022_ECCV}                                            &                               &                                                                            &                                                                        &                                                                      &                                                                           &                                                                         &                                                                       &                                                                     &                                                                                                        \\
            CODA-KITTI                                                                        & Cam., Lidar                 & 309                                                                        & Void Classes                                                           & Content                                                              & 6                                                                         & Bbox (2D)                                                     & \textbf{---}                                                          & \textbf{---}                                                        & \textbf{---}                                                          & KITTI$^\circledast$            \\
            CODA-nuScenes                                                                     & Cam., Lidar                 & 134                                                                        & Void Classes                                                           & Content                                                              & 17                                                                        & Bbox (2D)                                                     & \textbf{---}                                                          & \textbf{---}                                                        & \textbf{---}                                                          & nuScenes$^\circledast$         \\
            CODA-ONCE                                                                         & Cam., Lidar                 & 1,057                                                                      & OOD Detection                                                          & Content                                                              & 32                                                                        & Bbox (2D)                                                     & \textbf{---}                                                          & \textbf{---}                                                        & \textbf{---}                                                          & ONCE$^\circledast$             \\
            CODA2022-ONCE                                                                     & Cam., Lidar                 & 1,057                                                                  & OOD Detection                                                          & Content                                                              & 29                                                                        & Bbox (2D)                                                     & \textbf{---}                                                          & \textbf{---}                                                        & \cmark                                                                & ONCE$^\circledast$             \\
            CODA2022-SODA10M                                                                  & Cam.                        & 8,711                                                              & OOD Detection                                                          & Content                                                              & 29                                                                        & Bbox (2D)                                                     & \textbf{---}                                                          & \textbf{---}                                                        & \cmark                                                                & SODA10M$^\circledast$          \\
                      \midrule
            \textbf{W-OOD Tracking}~\cite{Maag_Video_2022_ACCV}                               &                               &                                                                            &                                                                        &                                                                      &                                                                           &                                                                         &                                                                       &                                                                     &                                                                       &                                \\
            Street Obstacle Sequences                                                         & Cam., Depth                 & \begin{tabular}[c]{@{}l@{}}1,129\end{tabular}                              & Recording                                                              & Content                                                              & 13                                                                        & Inst. mask (2D)                                                          & \cmark                                                                & \textbf{---}                                                        & \textbf{---}                                                          & Cityscapes$^\dagger$           \\
            CARLA-WildLife                                                                    & Cam., Depth                 & 1,210                                                                      & Simulation                                                             & Content                                                              & 18                                                                        & Inst. mask (2D)                                                          & \cmark                                                                & \textbf{---}                                                        & \cmark                                                                & \textbf{CARLA}                 \\ \midrule
            \textbf{Misc}                                                                     &                               &                                                                            &                                                                        &                                                                      &                                                                           &                                                                         &                                                                       &                                                                     &                                                                       &                                \\
            Lost and Found~\cite{Pinggera_Lost_2016_IROS}                                     & Stereo Cam.                & 2,104                                                                      & Recording                                                              & Content                                                              & 42                                                                        & Sem. mask (2D)                                                           & \cmark                                                                & \textbf{\textemdash}                                                & \textbf{---}                                                          & \textbf{---}                   \\
            WD-Pascal~\cite{Bevandic_Simultaneous_2019_GCPR}                                  & Cam.                        & 70                                                                         & Data Augmentation                                                      & Content                                                              & 1                                                                         & Sem. mask (2D)                                                          & \textbf{\textemdash}                                                  & \textbf{\textemdash}                                                & \textbf{\textemdash}                                                  & WildDash$^\circledast$         \\
            Vistas-NP~\cite{visapp21}                                                         & Cam.                        & 11,167                                                                     & Class Exclusion                                                        & Content                                                              & 4                                                                         & Sem. mask (2D)                                                          & \textbf{\textemdash}                                                  & \textbf{\textemdash}                                                & \cmark                                                                & Mapillary Vistas$^\circledast$ \\
            MUAD~\cite{franchi22bmvc}                                                         & Cam., Depth                 & 4,641                                                                      & Sim., Class Exclusion                                            & Content                                                              & 9                                                                         & Sem. mask (2D)                                                           & \textbf{\textemdash}                                                  & \textbf{\textemdash}                                                & \cmark                                                                & MUAD                           \\

            DeepAccident~\cite{wangDeepAccidentMotionAccident2023}                            & Cam., Lidar                 & 57,000                                                                     & Simulation                                                             & Temporal                                                             & 9                                                                         & Bbox (3D)                                                     & \cmark                                                                & \textbf{\textemdash}                                                & \cmark                                                                & \textbf{CARLA} 
            \\ \midrule

            \textbf{AnoVox (ours)}                                                            & \textbf{\begin{tabular}{@{}l@{}}Cam., Lidar \\ Depth\end{tabular}} & \textbf{245,600}                                                           & Simulation                                                             & \textbf{\begin{tabular}{@{}c@{}}Content \\ Temporal\end{tabular}}                                           & \textbf{178}                                                              & \textbf{\begin{tabular}{@{}c@{}}Inst. mask (2D,3D) \\ Voxel (3D)\end{tabular}}                                 & \cmark                                                                & \cmark                                                              & \cmark                                                                & \textbf{CARLA}
            \\ \bottomrule
        \end{tabular}%
    }
    \label{tab:overview}
\end{table*}

In autonomous driving, anomaly detection has been classically evaluated on a small set of benchmarks. For our analysis of related works, we have included works with published open-access perception datasets from an ego perspective that provide pixel- or point-wise ground truth. Thus, we neglected frameworks that do not provide explicit data~\cite{Xu_DANGER_2022_BayLearn, senaferreira:hal-03779723, hanselmann_king_2022, wiederer_benchmark_2022, Vadis_generating_2022_CVPR}, such with missing or incomplete data~\cite{gongSDACMultimodalSynthetic2024,liAutomatedEvaluationLarge2024,koselRevisitingOutofDistributionDetection2024}, and any that provide only frame-wise annotations~\cite{sakaridisACDCAdverseConditions2021, Zendel_2018_ECCV, tremblayRainRenderingEvaluating2021,aliakbarian2018viena2}. For our overview of related works as shown in Table~\ref{tab:overview}, we updated and extended the survey of Bogdoll et al.~\cite{bogdoll_perception}.


It can be observed that most benchmarks are small and designed for camera-based \textit{content} anomaly detection, providing ground truth in the form of semantic masks. Among the datasets including \textit{content} anomalies, just the CODA~\cite{Li_CODA_2022_ECCV} family includes lidar data but only provides ground truth in the form of 2D bounding boxes in the camera space. The DeepAccident~\cite{wangDeepAccidentMotionAccident2023} benchmark is the only one to provide \textit{temporal} 3D labels for lidar point clouds.

There are different categories of how anomalies were introduced~\cite{bogdoll_perception} in the datasets. \textbf{Recording} and \textbf{Simulation} are similar in the way that selected anomalies were directly introduced into the data. This way, the anomalies are truly part of the environment~\cite{Pinggera_Lost_2016_IROS, buCARLASimulatedData2021, Maag_Video_2022_ACCV}. The definition of what counts as an anomaly can vary, though. \textbf{Data Augmentation} typically follows a copy-and-paste pattern, where images of anomalies are pasted onto a scene from another dataset. This way, a distribution shift between the anomaly and the underlying data is introduced~\cite{Blum_Fishyscapes_2021_IJCV, Bevandic_Simultaneous_2019_GCPR}. \textbf{Web Sourcing} describes the process of manually curating images from the web that are deemed anomalous~\cite{Chan_SegmentMeIfYouCan_2021_NEURIPS,kimCrashNotCrash2019}. \textbf{Class Exclusion} is based on existing datasets and removes selected classes from the training data, thus treating them as anomalous, while the classes themselves remain rather normal from a human point of view~\cite{grcic_dense_2021,Hendrycks_Scaling_2022_ICML}. \textbf{Void Classes} utilizes void or misc classes from existing datasets and labels them as anomalous. This can be done with additional labeling guidelines~\cite{Li_CODA_2022_ECCV}. Finally, \textbf{OOD Detection} uses an anomaly detection method to derive anomaly proposals from a dataset which can then be labeled, typically after a human quality inspection~\cite{Li_CODA_2022_ECCV}.

\begin{figure}[t!]
  \centering
  \includegraphics[width=\textwidth]{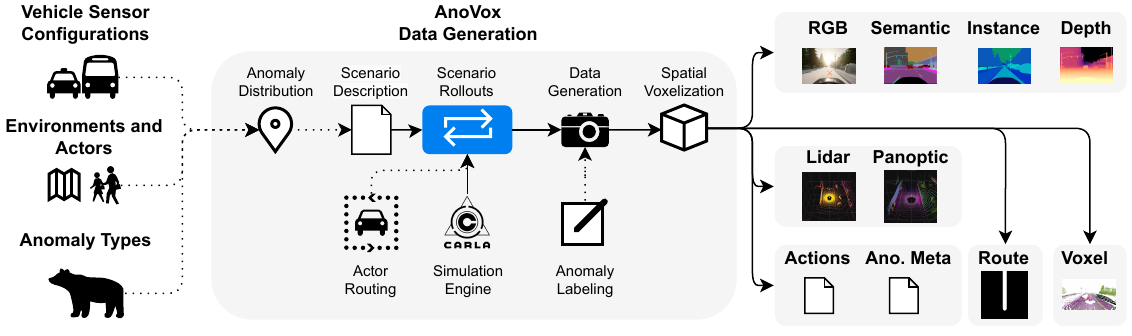}
  \caption{\textbf{Overview:} With Anovox, highly configurable scenarios can be created that either represent \textit{normality} or include \textit{content} or \textit{temporal} anomalies. Generated datasets include rich labels and ground truth in 2D, 3D, and a spatial voxel space.}
  \label{fig:overview}
\end{figure}

While many datasets include temporal data in the form of sequences and provide labels for regular tasks, such as object detection or semantic segmentation, none includes state information about the ego vehicle. However, during deployment, such information is generally available and can be leveraged.

Most benchmarks do not provide a definition of normality~\cite{Blum_Fishyscapes_2021_IJCV,Chan_SegmentMeIfYouCan_2021_NEURIPS,Hendrycks_Scaling_2022_ICML}. This makes it especially hard for un- and self-supervised methods to precisely detect anomalies if the semantic training distribution is not fully known. Especially desirable is a well-defined normality that allows for the generation of compliant training data. This is only possible in simulation, where full control over both the training and evaluation data is available. Sometimes, even unlabeled anomalies occur in evaluation data~\cite{wangDeepAccidentMotionAccident2023}. 

AnoVox is a challenging benchmark that addresses anomaly detection from an embodied AI perspective rather than from a pure computer vision perspective. We provide data in a temporal context with state information about the ego vehicle for typical multimodal sensor setups. For a clear definition of what anomalies are, we provide a formalized definition of normality and compliant training data. This allows for the usage of un- and self-supervised anomaly detection methods, which often rely on large amounts of unlabeled data.

%% file: sec/3_data.tex
\section{AnoVox Dataset}
\label{sec:anovox}

\begin{table}[t]
\centering
\caption{A formal definition of normality and how the training data generated and provided by AnoVox aligns with the definition.}
\label{tab:normality}
\resizebox{0.7\textwidth}{!}{%
\begin{tabular}{@{}lll@{}}
\toprule
          \textbf{Category}                 & \textbf{Description}            & \textbf{AnoVox}                                                                                                                                                                                                    \\ \midrule
\textbf{Ego}               &                                 &                                                                                                                                                                                                                                 \\
Ego vehicle                & Recording vehicle               & Lincoln MKZ 2020                                                                                                                                                                                                                \\
Sensor config.       & Sensor types, placements & Configurations Mono, Stereo, Multi, Surround                                                                                                                                                                                             \\
Ego behavior               & Driving characteristics         & Behavior Agent (actions)                                                                                                                                                                                                                  \\ \midrule
\textbf{Domain}            &                                 &                                                                                                                                                                                                                                 \\
Area                       & Geographical area               & Towns 01,02,03,04,05,06,07,10HD                                                                                                                                                                                                 \\
Environment                & Weather, time of day            & \begin{tabular}[c]{@{}l@{}}ClearNoon, CloudyNoon, WetNoon,\\ WetCloudyNoon, HardRainNoon,\\ SoftRainNoon, ClearSunset, CloudySunset,\\ WetSunset, WetCloudySunset, MidRainSunset,\\ HardRainSunset, SoftRainSunset\end{tabular} \\ \midrule
\textbf{Physical entities} &                                 &                                                                                                                                                                                                                                 \\
Traffic participants       & Vehicles, VRUs                  & Vehicles, Walkers $\in$ Blueprint library                                                                                                                                                                                       \\
Vehicle behavior           & Driving characteristics         & Traffic Manager Autopilot                                                                                                                                                                                       \\
Pedestrian behavior        & Movement characteristics        & AI Walker \\\bottomrule                  
\end{tabular}%
}
\end{table}

AnoVox provides both data that represents \textit{normality} and data including \textit{content} and \textit{temporal} anomalies. While we do provide a large-scale dataset, AnoVox is primarily a scalable framework that can be used with arbitrary vehicle setups and a wide selection of parameters to create further data. This does not only allow for the detection of anomalies in known environments but also for the detection of anomalies under domain shifts. In Sec.~\ref{subsec:normality}, we provide a formalized definition of normality and demonstrate how the training data provided by AnoVox follows this definition of normality. In Sec.~\ref{subsec:scenarios}, we show what types of scenarios can be generated with AnoVox. Finally, Sec.~\ref{subsec:dataset} provides an overview of our dataset.

\subsection{Definition of Normality}
\label{subsec:normality}

In the past, the definition of normality was often not extensively discussed when anomaly detection benchmarks were presented. A typical solution is to define normality as all semantic classes from the Cityscapes dataset~\cite{Blum_Fishyscapes_2021_IJCV,Chan_SegmentMeIfYouCan_2021_NEURIPS}. However, this is not necessarily related to the training data. And if it is, it requires labeled training data as normality to be aware of the classes.

With AnoVox, we provide full controllability about both normality and anomalies in synthetic environments. This ensures that anomalies included in the benchmark are true anomalies and are not included in an unlabeled training dataset by chance. For a fair benchmark of anomaly detection methods, it is important that they share the same definition of normality. We argue that this normality needs to be primarily defined by the training data rather than expert-defined concepts. This, however, requires the option to generate large amounts of training data following a definition of normality, which is infeasible in the real world, as anomalies would certainly occur in fleet-sized, unlabeled datasets. However, in the field of autonomous driving as a subfield of embodied AI, there is more to the training data than just frames: There is a recording entity that performs actions, and there is temporal context. Therefore, we provide a formal definition of normality based on three categories:

\textbf{Ego:} Domain shifts in data cannot only be induced by novel environmental conditions but also by different capturing methods. In autonomous driving, this especially refers to sensor types and configurations. In addition, temporal changes in the environment are heavily influenced by our own actions. Thus, also the behavior of the ego agent counts towards normality.

\textbf{Domain:} With the domain, we describe the static environment around the vehicle. This includes the geographical areas the vehicle has traversed, but also seen weather types and time of day specifications.

\textbf{Physical Entities:} These are the dynamic actors in the scene, most typically other vehicles, cyclists, and pedestrians. However, also categories such as animals or potentially moving objects can be included here.

Based on these concepts, we provide a definition of normality for the CARLA simulation engine in Table~\ref{tab:normality} that allows for the generation of compliant datasets.

\begin{figure}[t]
    \centering
        \resizebox{.8\textwidth}{!}{\input{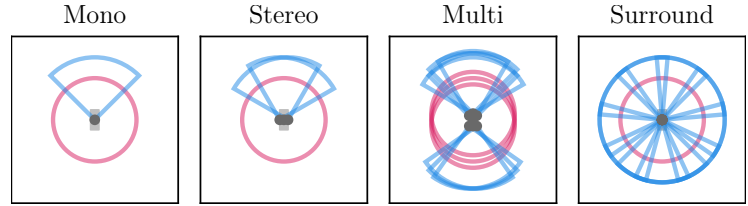}}
    \caption{Preconfigured sensor configurations in the AnoVox suite. Blue wedges visualize RGB cameras, while red circles visualize lidar sensors.}
    \label{fig:sensors}
\end{figure}

\subsection{Scenario Generation}
\label{subsec:scenarios}

AnoVox is designed to create configurable, large-scale datasets that either define normality or include anomalies. As shown in Fig.~\ref{fig:overview}, first, the vehicle sensor configuration needs to be set. AnoVox currently supports an arbitrary number of camera, lidar, and depth sensors, which can be positioned freely on the ego vehicle of choice. This allows for the replication of existing sensor setups, the alignment with other datasets, or the testing of new configurations.

\begin{figure}[t!]
  \centering
  \begin{tabular}{ccc}
    \includegraphics[width=0.32\textwidth]{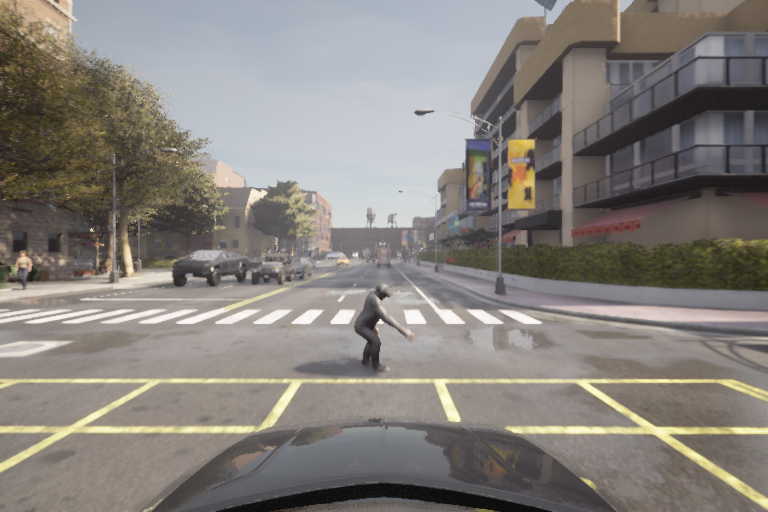} &
    \includegraphics[width=0.32\textwidth]{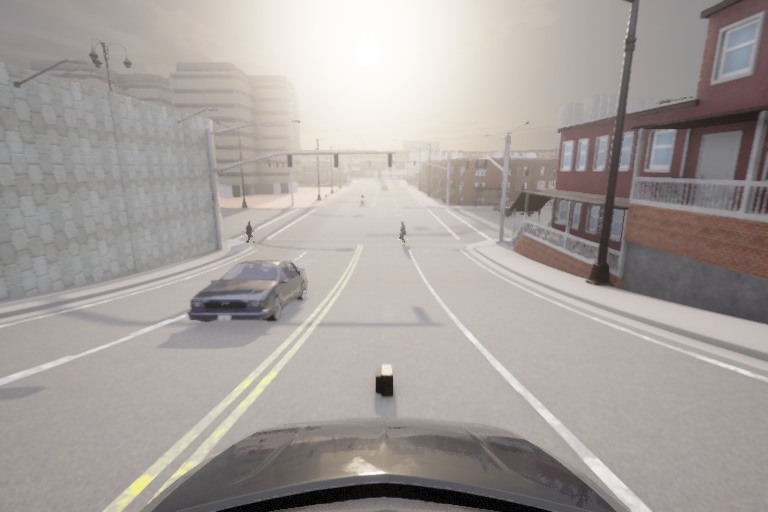} &
    \includegraphics[width=0.32\textwidth] {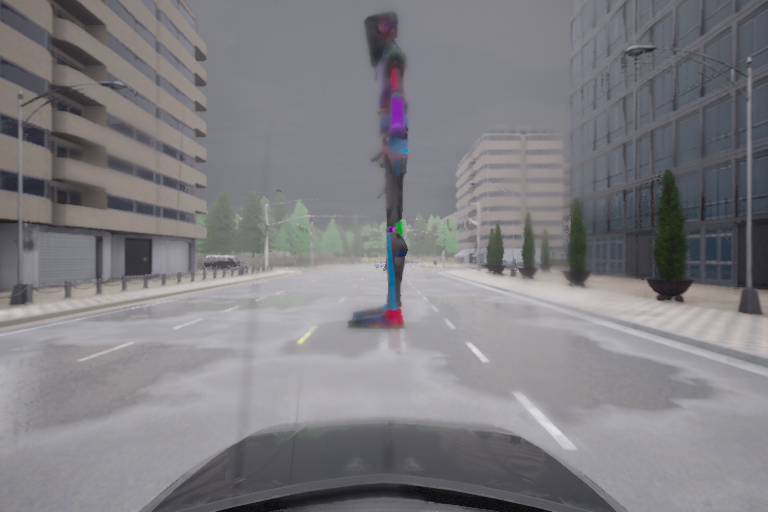} \\
    \includegraphics[width=0.32\textwidth]{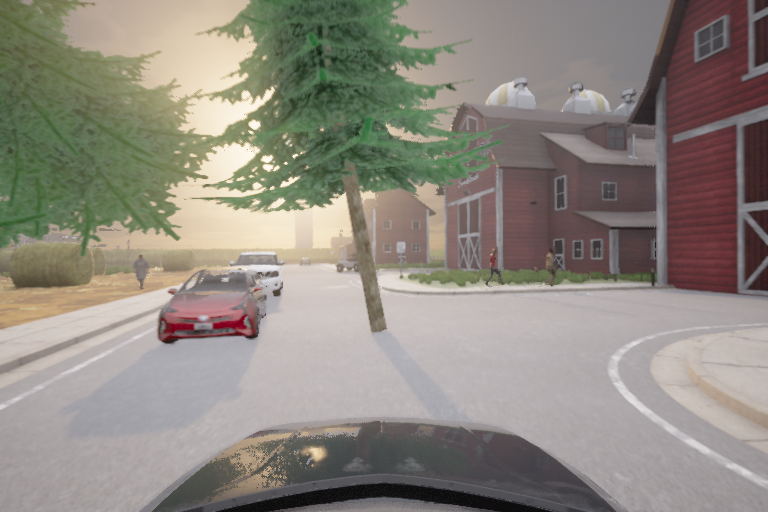} & 
    \includegraphics[width=0.32\textwidth]{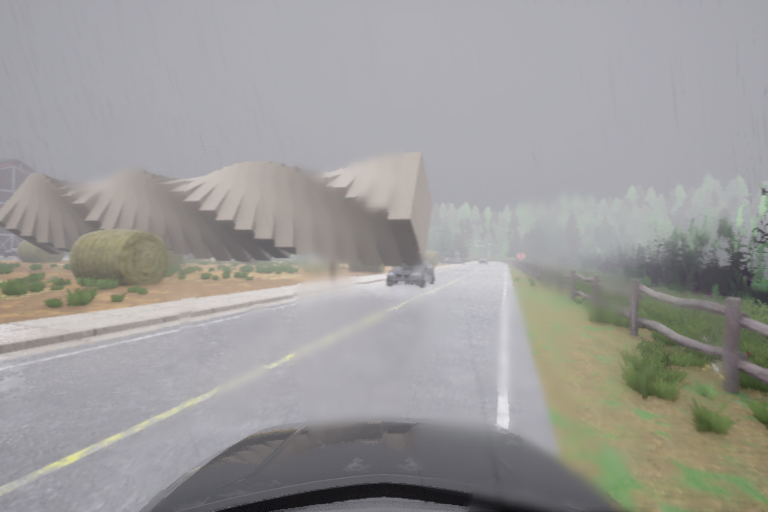} & 
    \includegraphics[width=0.32\textwidth]{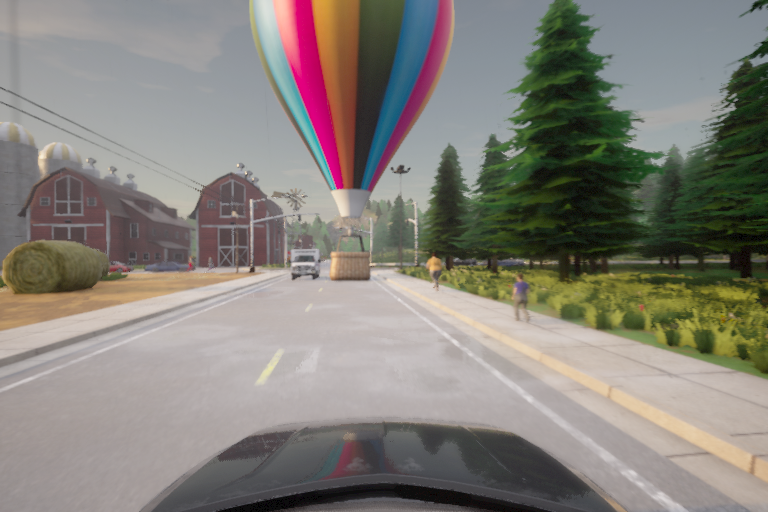}
  \end{tabular}
  \caption{\textbf{Content anomalies:} Examples from our six categories: An ape as an animal anomaly, an old tv as a home anomaly, a statue as a special anomaly, a tree as a nature anomaly, a pillar as a falling anomaly, and a hot air balloon as an airplane anomaly.}
  \label{fig:content_ano}
\end{figure}

We provide four pre-designed multimodal vehicle sensor configurations. As shown in Fig.~\ref{fig:sensors}, the \textbf{mono} configuration consists of a lidar and a camera which are centered on top of the vehicle, centered forward. Our \textbf{stereo} setup consists of two cameras at the front edge and both a camera and lidar on top. The \textbf{multi} setup adds rear-facing cameras and additional lidar sensors at the front and rear, positioned at a lower level compared to the roof-mounted lidars. Finally, the \textbf{surround} setup provides a full 360° camera view next to a top-mounted lidar. Every RGB camera automatically comes with a depth camera.

Second, the environment and actors need to be set. AnoVox currently supports eight different regions and 14 weather and time of day presets. Pedestrians, cyclists, as well as multiple types of vehicles can be spawned in the scene.

Third, the type of anomaly needs to be defined. Currently, AnoVox supports to create \textit{normality} without anomalies, the placement of \textit{content} anomalies, and the activation of \textit{temporal} anomalies. By removing domains from the training dataset, AnoVox also supports the detection of anomalies under domain shifts.

Given these configurations, AnoVox pre-computes the scenario flows and stores them as scenario descriptions. Thus, meta-data describing all scenarios is available and allows for effortless dataset analysis. Scenario descriptions can be shared among peers by sharing the scenario description files and locally generating the scenarios without the need to move hundreds of gigabytes of data.

Based on the given configuration, AnoVox executes the driving scenarios in simulation. We provide a custom-built CARLA simulation engine~\cite{Dosovitskiy2017CARLAAO} including all \textit{content} anomalies that we have manually collected and processed. Each scenario has a length of 20~seconds and is recorded at 10~Hz, resulting in 200 frames. When the scenario starts, the ego vehicle is spawned in the world and follows a given route to its target goal. At some point along the way, a \textit{content} or \textit{temporal} anomaly will appear. To guarantee that the anomaly is reached in time, a green wave is activated along the route of the ego vehicle. Since physics computations remain active in the simulation, the ego vehicle will make contact with the anomalies, which leads to realistic collisions.

\subsubsection{Actor Routing.}
\label{subsec:actor_routing}
Strange objects on the road do the same thing in simulation as they would do in real life - they cause traffic jams. This makes it often unfeasible to reach the anomaly for the ego vehicle. Thus, we deploy a filter and rerouting algorithm for all other vehicles in our scenario. First, we filter all actors close to our spawn point and on our direct path toward the anomaly. Then, we continuously monitor all planned paths and reroute vehicles whenever they were to enter a lane with an anomaly on it, as they would get stuck. This rerouting technically changes their driving behavior. As shown in Table~\ref{tab:normality}, in the training data, only vehicles are present, which show an \textit{autopilot} driving behavior. Rerouting makes them switch into a \textit{behavior agent}. We deal with this by providing novel labels, as further explained in Sec.~\ref{subsec:label}. This way, false positives, which might occur due to a violation of the alignment with normality, can be filtered out for evaluation.

\begin{figure}[t!]
    \centering
    \begin{subfigure}{0.24\textwidth}
        \includegraphics[width=\textwidth]{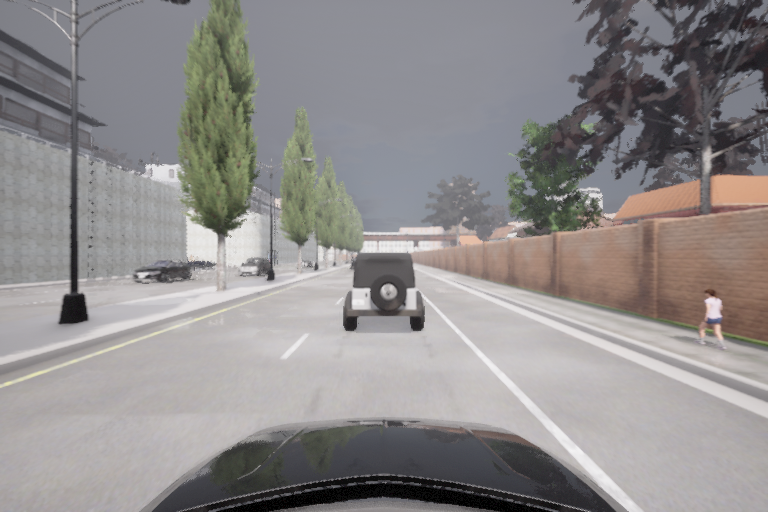}
    \end{subfigure}
    \begin{subfigure}{0.24\textwidth}
        \includegraphics[width=\textwidth]{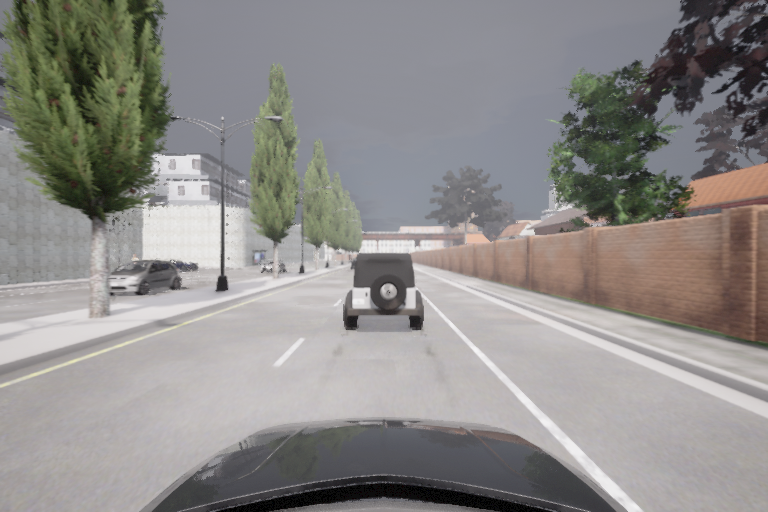}
    \end{subfigure}
    \begin{subfigure}{0.24\textwidth}
        \includegraphics[width=\textwidth]{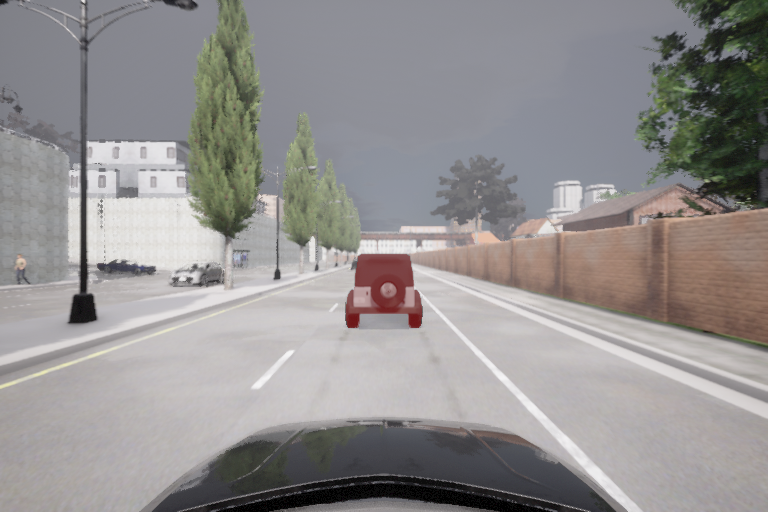}
    \end{subfigure}
    \begin{subfigure}{0.24\textwidth}
        \includegraphics[width=\textwidth]{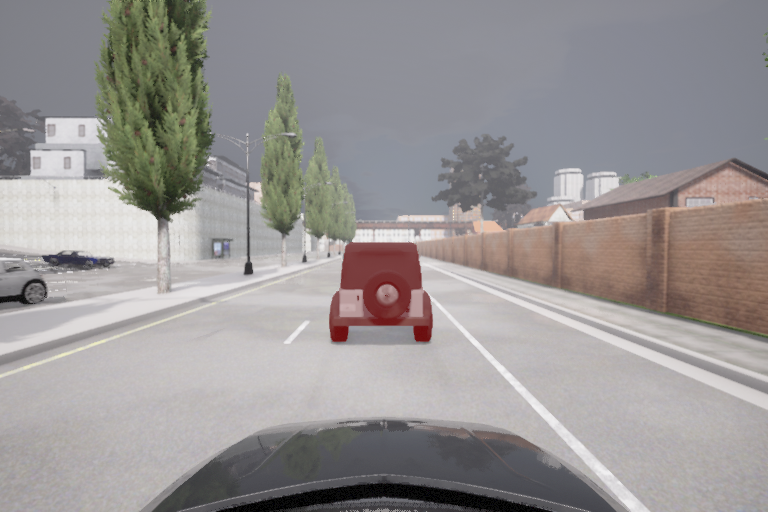}
    \end{subfigure}
    \caption{\textbf{Temporal anomalies:} This scenario shows our implemented type of \textit{temporal} anomalies. The first two images show the regular vehicle following mode. The third and fourth images show the active braking maneuver with overlayed ground truth.}
    \label{fig:temporal_ano}
\end{figure}

\subsubsection{Content Anomalies.}
\label{subsec:scenarios_content}

We provide 178 different \textit{content} anomalies in five different size classes \textit{tiny, small, medium, big}, and \textit{huge}. Semantically, they hierarchically belong to six different super-classes. Every anomaly has an individual label next to its super-class and size label. The class \textbf{animal} includes 33 animals of different sizes.  The category \textbf{home} includes 53 typical household items such as furniture, tables, backpacks, or cardboard boxes. The category \textbf{special} includes 67 objects of rather atypical types and such that fall into a misc category, such as Pokemon, some of which could appear in the real world in the form of costumes or cuddly toys. The class \textbf{nature} includes 12 outdoor objects, such as rocks or wood. In the category \textbf{falling}, we provide 9 large objects, such as novel trees, that were spawned in an unstable position, which made them fall over. Finally, the \textbf{airplane} class consists of four types of large flying objects. 

As especially large anomalies sometimes require manual positioning, we use the classes \textit{home, special,} and \textit{animal} for our automated scenario generation. Here, we place all anomalies in critical positions along the way distributed around lane centers. For the classes \textit{nature, falling,} and \textit{airplane}, we provide a manually curated dataset in a geographic area with large free-space areas.

\subsubsection{Temporal Anomalies.}
\label{subsec:scenarios_temporal}

There are two ways to generate \textit{temporal} anomalies: Optimization-based or knowledge-based~\cite{wangDeepAccidentMotionAccident2023}. While both are of value, for our benchmark, it is important to comply with the defined normality, which is more challenging with optimization-based methods. Thus, we provide knowledge-based \textit{content} anomalies. As prior work has shown that such expert-defined temporal anomalies require high engineering efforts~\cite{Bogdoll_Ontology_2022_ECCV, Bogdoll_Quantification_2022_IV,aliakbarian2018viena2}, we provide only a single type of \textit{temporal} anomalies. We implement \textbf{sudden braking} scenarios of a lead vehicle, which is a very typical scenario in everyday traffic~\cite{kimCrashNotCrash2019}. While we set a planned route for the ego vehicle in our \textit{content} anomaly scenarios, we use the same route and follow it with a lead vehicle. Then, our ego vehicle follows that lead vehicle. Along the route, the lead vehicle will perform a sudden brake with negative velocities which cannot be seen during training. While braking, we label the lead vehicle in the same way as we label \textit{content} anomalies.


\begin{figure}[tbp]
\centering
    \resizebox{0.6\textwidth}{!}{
    \begin{subfigure}{0.48\textwidth}
        \includegraphics[width=\textwidth]{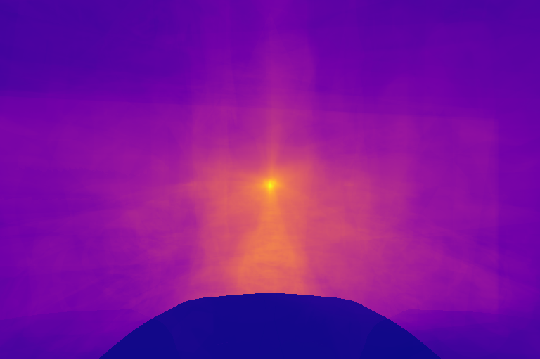}
    \end{subfigure}
    \begin{subfigure}{0.48\textwidth}
        \includegraphics[width=\textwidth]{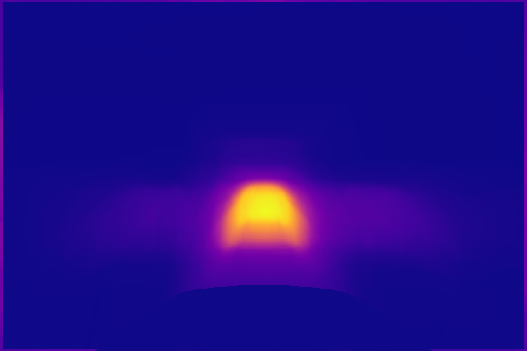}
    \end{subfigure}}
    \caption{Distribution of our \textit{content} anomalies (left) and distribution of the \textit{temporal} anomalies (right), based on the 2D semantic labels in the dataset.}
  \label{fig:heat_anomalies}
\end{figure}

\subsubsection{Data Generation.}
\label{subsec:data_gen}

As shown in Fig.~\ref{fig:overview}, we provide sensor data for all positioned RGB cameras, depth cameras, and lidar sensors. For regular perception tasks, we provide panoptic masks for both camera and lidar. For each frame, also information about the state of the ego vehicle is collected, such as its actions \textit{throttle, street, and brake}. For approaches that require additional information about the planned route~\cite{zhang2021roach, chauffeur,mile2022,bogdoll2023muvo}, we provide a standard-format Bird's-eye view~(BEV) representation of the planned route. 

For the anomalous instances, we provide meta-data, such as their positions and their size. Ground truth is embedded into the semantic masks for both camera and lidar. As we designed the benchmark to compare methods that use different modalities for the detection of anomalies, we represent all anomalies also in a 3D voxel grid with customizable grid size. Based on our depth maps and lidar point clouds, we fuse all visible points from all sensors in 3D and quantize them into the voxel grid. In addition, the task of detecting anomalies in 3D is especially helpful for downstream tasks, such as motion planning.

By using the CARLA simulation engine, we support the largest simulation ecosystem in autonomous driving. However, it has inherent flaws, such as imperfect behavior agents~\cite{carla_issues}. Thus, we perform minor data cleaning to remove scenarios that contain collisions with pedestrians in the evaluation data and scenarios that show high deceleration values for other vehicles in the training data. This ensures that we avoid unlabeled anomalies in the evaluation data and that our temporal anomalies are distinct from all behaviors seen during training.

\subsubsection{Labeling.}
\label{subsec:label}

\begin{figure}[tbp]
\begin{Large}

\centering
\resizebox{0.9\textwidth}{!}{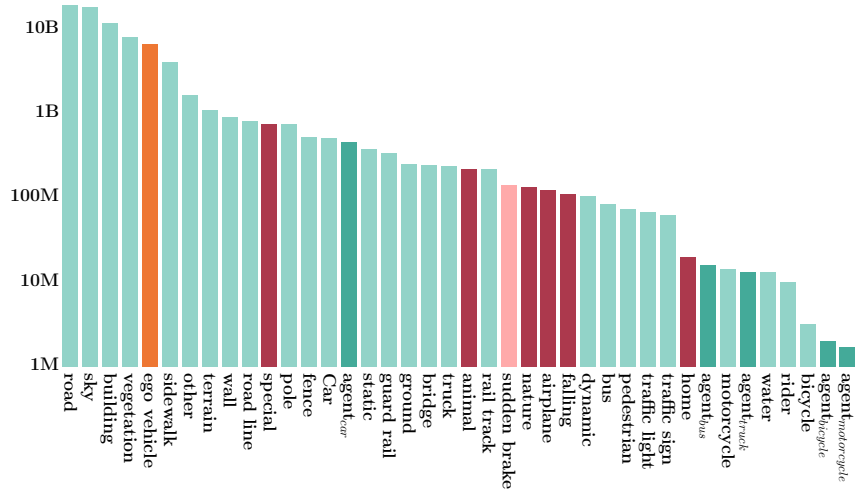}
\vspace*{4mm}
\caption{\textbf{Number of pixels per class:} Bright green represents standard classes, and dark green ones additional classes for vehicles that switched into an agent behavior mode. The orange class shows our ego vehicle, and all red classes represent anomalies.}
  \label{fig:class_labels}
  \end{Large}
\end{figure}

AnoVox provides 40 label classes, as shown in Fig.~\ref{fig:class_labels}. We provide labels for standard tasks as used by Cityscapes and CARLA~\cite{cordts_cityscapes_2016,Dosovitskiy2017CARLAAO}. Additionally, we label the ego vehicle if visible. As described in Sec.~\ref{subsec:actor_routing}, some vehicles in the scene might switch from autopilot to a behavior agent. As this driving behavior is not present in training data, we provide additional labels for all vehicle classes while controlled by a behavior agent. Finally, we provide labels for all \textit{content} and \textit{temporal} anomalies. Labels for the super-classes can be found in the semantic labels, while fine-granular, individual anomaly labels are provided in additional metadata. Finally, voxels are provided with an anomaly label when the closest point to their center is anomalous.



\subsection{Dataset and Statistics}
\label{subsec:dataset}

Next to all code to generate further datasets, we provide over 1.3~TB of generated data. We have split the data into chunks smaller than 50~GB for easier processing, which we call instantiations. We provide 1,117 scenarios based on the \textit{mono} sensor configuration, 76 for the \textit{stereo} configuration, and 35 for the \textit{multi} configuration. While the \textit{stereo} and \textit{multi} configurations are more of a symbolic nature, our dataset for the \textit{mono} sensor configuration is extensive. We provide data for eight different areas, and for each, we record \textit{normality} training data as well as evaluation data with \textit{content} and \textit{temporal} anomalies, resulting in 24 different types of scenarios. Additionally, for each of those 24, we create many scenarios where settings such as the weather, time of day, or spawned anomaly vary. For \textit{content} anomalies, 14.8\% of all frames contain visible anomalies in camera data and 74.8\% in lidar data. For \textit{temporal} anomalies, 15.5\% of all frames include anomalies equally visible in both sensor modalities. As shown in Fig.~\ref{fig:heat_anomalies} anomalies are well distributed over the visible space but with a focus on critical areas in front of the ego vehicle.


Finally, Fig.~\ref{fig:class_labels} gives an overview of the number of pixels per class label in our dataset. We support all basic classes and provide additional classes for our ego vehicle and other traffic participants that have switched from autopilot to an agent behavior, so they can be excluded from evaluation if needed, as they behave slightly differently than the ones available in the training data.

\begin{table}[t]
\centering
\caption{A comparison of current state-of-the-art methods for anomaly detection.}
\label{tab:sota_anodec}
\resizebox{0.5\textwidth}{!}{%
\begin{tabular}{@{}llcc@{}}
\toprule
                                                                     & \textbf{Year} & \textbf{Sensor} & \textbf{Aux}   \\ \midrule
3DUIS~\cite{nunes_unsupervised_2022}               & 2022          & LiDAR             & \xmark \\
APF~\cite{li_open-set_2023}                        & 2023          & LiDAR             & \xmark \\
cDNP~\cite{galesso_far_2023}                       & 2023          & Camera            & \xmark \\
EAM~\cite{grcic_advantages_2023}                   & 2023          & Camera            & \cmark \\
ElC-OIS ~\cite{deng_elc-ois_2023}                  & 2023          & LiDAR             & \xmark \\
LS-VOS~\cite{piroli_ls-vos_2023}                   & 2023          & LiDAR             & \xmark \\
Mask2Anomaly~\cite{rai_unmasking_2023}             & 2023          & Camera            & \cmark \\
Maskomaly~\cite{ackermann_maskomalyzero-shot_2023} & 2023          & Camera            & \xmark \\
Maximized Entropy~\cite{chan_entropy_2021}         & 2021          & Camera            & \cmark \\
Najibi et al.~\cite{najibi_motion_2022}            & 2022          & LiDAR             & \xmark \\
ObsNet~\cite{besnier_triggering_2021}              & 2021          & Camera            & \xmark \\
RbA~\cite{nayal_rba_2023}                          & 2023          & Camera            & Both                  \\
ReaL~\cite{cen_open-world_2022}                    & 2022          & LiDAR             & Both                  \\
SynBoost~\cite{di_biase_pixel-wise_2021}          & 2021          & Camera            & \cmark \\ \bottomrule
\end{tabular}%
}
\end{table}

%% file: img/anovox_classes.pdf_tex
\begingroup%
  \makeatletter%
  \providecommand\color[2][]{%
    \errmessage{(Inkscape) Color is used for the text in Inkscape, but the package 'color.sty' is not loaded}%
    \renewcommand\color[2][]{}%
  }%
  \providecommand\transparent[1]{%
    \errmessage{(Inkscape) Transparency is used (non-zero) for the text in Inkscape, but the package 'transparent.sty' is not loaded}%
    \renewcommand\transparent[1]{}%
  }%
  \providecommand\rotatebox[2]{#2}%
  \newcommand*\fsize{\dimexpr\f@size pt\relax}%
  \newcommand*\lineheight[1]{\fontsize{\fsize}{#1\fsize}\selectfont}%
  \ifx\svgwidth\undefined%
    \setlength{\unitlength}{652.70333862bp}%
    \ifx\svgscale\undefined%
      \relax%
    \else%
      \setlength{\unitlength}{\unitlength * \real{\svgscale}}%
    \fi%
  \else%
    \setlength{\unitlength}{\svgwidth}%
  \fi%
  \global\let\svgwidth\undefined%
  \global\let\svgscale\undefined%
  \makeatother%
  \begin{picture}(1,0.58600495)%
    \lineheight{1}%
    \setlength\tabcolsep{0pt}%
    \put(0,0){\includegraphics[width=\unitlength,page=1]{anovox_classes.pdf}}%
    \put(0.04236736,0.14022433){\color[rgb]{0,0,0}\rotatebox{-90}{\makebox(0,0)[lt]{\lineheight{1.25}\smash{\begin{tabular}[t]{l}\textbf{road}\end{tabular}}}}}%
    \put(0.06649778,0.14022433){\color[rgb]{0,0,0}\rotatebox{-90}{\makebox(0,0)[lt]{\lineheight{1.25}\smash{\begin{tabular}[t]{l}\textbf{sky}\end{tabular}}}}}%
    \put(0.09062819,0.14022433){\color[rgb]{0,0,0}\rotatebox{-90}{\makebox(0,0)[lt]{\lineheight{1.25}\smash{\begin{tabular}[t]{l}\textbf{building}\end{tabular}}}}}%
    \put(0.1147586,0.14022433){\color[rgb]{0,0,0}\rotatebox{-90}{\makebox(0,0)[lt]{\lineheight{1.25}\smash{\begin{tabular}[t]{l}\textbf{vegetation}\end{tabular}}}}}%
    \put(0.13888901,0.14022433){\color[rgb]{0,0,0}\rotatebox{-90}{\makebox(0,0)[lt]{\lineheight{1.25}\smash{\begin{tabular}[t]{l}\textbf{ego vehicle}\end{tabular}}}}}%
    \put(0.16301942,0.14022433){\color[rgb]{0,0,0}\rotatebox{-90}{\makebox(0,0)[lt]{\lineheight{1.25}\smash{\begin{tabular}[t]{l}\textbf{sidewalk}\end{tabular}}}}}%
    \put(0.18714983,0.14022433){\color[rgb]{0,0,0}\rotatebox{-90}{\makebox(0,0)[lt]{\lineheight{1.25}\smash{\begin{tabular}[t]{l}\textbf{other}\end{tabular}}}}}%
    \put(0.21128024,0.14022433){\color[rgb]{0,0,0}\rotatebox{-90}{\makebox(0,0)[lt]{\lineheight{1.25}\smash{\begin{tabular}[t]{l}\textbf{terrain}\end{tabular}}}}}%
    \put(0.23541066,0.14022433){\color[rgb]{0,0,0}\rotatebox{-90}{\makebox(0,0)[lt]{\lineheight{1.25}\smash{\begin{tabular}[t]{l}\textbf{wall}\end{tabular}}}}}%
    \put(0.25954107,0.14022433){\color[rgb]{0,0,0}\rotatebox{-90}{\makebox(0,0)[lt]{\lineheight{1.25}\smash{\begin{tabular}[t]{l}\textbf{road line}\end{tabular}}}}}%
    \put(0.28367148,0.14022433){\color[rgb]{0,0,0}\rotatebox{-90}{\makebox(0,0)[lt]{\lineheight{1.25}\smash{\begin{tabular}[t]{l}\textbf{special}\end{tabular}}}}}%
    \put(0.30780189,0.14022433){\color[rgb]{0,0,0}\rotatebox{-90}{\makebox(0,0)[lt]{\lineheight{1.25}\smash{\begin{tabular}[t]{l}\textbf{pole}\end{tabular}}}}}%
    \put(0.3319323,0.14022433){\color[rgb]{0,0,0}\rotatebox{-90}{\makebox(0,0)[lt]{\lineheight{1.25}\smash{\begin{tabular}[t]{l}\textbf{fence}\end{tabular}}}}}%
    \put(0.35606271,0.14022433){\color[rgb]{0,0,0}\rotatebox{-90}{\makebox(0,0)[lt]{\lineheight{1.25}\smash{\begin{tabular}[t]{l}\textbf{Car}\end{tabular}}}}}%
    \put(0.38019312,0.14022433){\color[rgb]{0,0,0}\rotatebox{-90}{\makebox(0,0)[lt]{\lineheight{1.25}\smash{\begin{tabular}[t]{l}\textbf{agent$_{car}$}\end{tabular}}}}}%
    \put(0.40432353,0.14022433){\color[rgb]{0,0,0}\rotatebox{-90}{\makebox(0,0)[lt]{\lineheight{1.25}\smash{\begin{tabular}[t]{l}\textbf{static}\end{tabular}}}}}%
    \put(0.42845395,0.14022433){\color[rgb]{0,0,0}\rotatebox{-90}{\makebox(0,0)[lt]{\lineheight{1.25}\smash{\begin{tabular}[t]{l}\textbf{guard rail}\end{tabular}}}}}%
    \put(0.45258436,0.14022433){\color[rgb]{0,0,0}\rotatebox{-90}{\makebox(0,0)[lt]{\lineheight{1.25}\smash{\begin{tabular}[t]{l}\textbf{ground}\end{tabular}}}}}%
    \put(0.47671477,0.14022433){\color[rgb]{0,0,0}\rotatebox{-90}{\makebox(0,0)[lt]{\lineheight{1.25}\smash{\begin{tabular}[t]{l}\textbf{bridge}\end{tabular}}}}}%
    \put(0.50084518,0.14022433){\color[rgb]{0,0,0}\rotatebox{-90}{\makebox(0,0)[lt]{\lineheight{1.25}\smash{\begin{tabular}[t]{l}\textbf{truck}\end{tabular}}}}}%
    \put(0.52497559,0.14022433){\color[rgb]{0,0,0}\rotatebox{-90}{\makebox(0,0)[lt]{\lineheight{1.25}\smash{\begin{tabular}[t]{l}\textbf{animal}\end{tabular}}}}}%
    \put(0.549106,0.14022433){\color[rgb]{0,0,0}\rotatebox{-90}{\makebox(0,0)[lt]{\lineheight{1.25}\smash{\begin{tabular}[t]{l}\textbf{rail track}\end{tabular}}}}}%
    \put(0.57323641,0.14022433){\color[rgb]{0,0,0}\rotatebox{-90}{\makebox(0,0)[lt]{\lineheight{1.25}\smash{\begin{tabular}[t]{l}\textbf{sudden brake}\end{tabular}}}}}%
    \put(0.59736682,0.14022433){\color[rgb]{0,0,0}\rotatebox{-90}{\makebox(0,0)[lt]{\lineheight{1.25}\smash{\begin{tabular}[t]{l}\textbf{nature}\end{tabular}}}}}%
    \put(0.62149724,0.14022433){\color[rgb]{0,0,0}\rotatebox{-90}{\makebox(0,0)[lt]{\lineheight{1.25}\smash{\begin{tabular}[t]{l}\textbf{airplane}\end{tabular}}}}}%
    \put(0.64562765,0.14022433){\color[rgb]{0,0,0}\rotatebox{-90}{\makebox(0,0)[lt]{\lineheight{1.25}\smash{\begin{tabular}[t]{l}\textbf{falling}\end{tabular}}}}}%
    \put(0.66975806,0.14022433){\color[rgb]{0,0,0}\rotatebox{-90}{\makebox(0,0)[lt]{\lineheight{1.25}\smash{\begin{tabular}[t]{l}\textbf{dynamic}\end{tabular}}}}}%
    \put(0.69388847,0.14022433){\color[rgb]{0,0,0}\rotatebox{-90}{\makebox(0,0)[lt]{\lineheight{1.25}\smash{\begin{tabular}[t]{l}\textbf{bus}\end{tabular}}}}}%
    \put(0.71801888,0.14022433){\color[rgb]{0,0,0}\rotatebox{-90}{\makebox(0,0)[lt]{\lineheight{1.25}\smash{\begin{tabular}[t]{l}\textbf{pedestrian}\end{tabular}}}}}%
    \put(0.74214929,0.14022433){\color[rgb]{0,0,0}\rotatebox{-90}{\makebox(0,0)[lt]{\lineheight{1.25}\smash{\begin{tabular}[t]{l}\textbf{traffic light}\end{tabular}}}}}%
    \put(0.7662797,0.14022433){\color[rgb]{0,0,0}\rotatebox{-90}{\makebox(0,0)[lt]{\lineheight{1.25}\smash{\begin{tabular}[t]{l}\textbf{traffic sign }\end{tabular}}}}}%
    \put(0.79041011,0.14022433){\color[rgb]{0,0,0}\rotatebox{-90}{\makebox(0,0)[lt]{\lineheight{1.25}\smash{\begin{tabular}[t]{l}\textbf{home}\end{tabular}}}}}%
    \put(0.81454053,0.14022433){\color[rgb]{0,0,0}\rotatebox{-90}{\makebox(0,0)[lt]{\lineheight{1.25}\smash{\begin{tabular}[t]{l}\textbf{agent$_{bus}$}\end{tabular}}}}}%
    \put(0.83867094,0.14022433){\color[rgb]{0,0,0}\rotatebox{-90}{\makebox(0,0)[lt]{\lineheight{1.25}\smash{\begin{tabular}[t]{l}\textbf{motorcycle}\end{tabular}}}}}%
    \put(0.86280135,0.14022433){\color[rgb]{0,0,0}\rotatebox{-90}{\makebox(0,0)[lt]{\lineheight{1.25}\smash{\begin{tabular}[t]{l}\textbf{agent$_{truck}$}\end{tabular}}}}}%
    \put(0.88693176,0.14022433){\color[rgb]{0,0,0}\rotatebox{-90}{\makebox(0,0)[lt]{\lineheight{1.25}\smash{\begin{tabular}[t]{l}\textbf{water}\end{tabular}}}}}%
    \put(0.91106217,0.14022433){\color[rgb]{0,0,0}\rotatebox{-90}{\makebox(0,0)[lt]{\lineheight{1.25}\smash{\begin{tabular}[t]{l}\textbf{rider}\end{tabular}}}}}%
    \put(0.93519258,0.14022433){\color[rgb]{0,0,0}\rotatebox{-90}{\makebox(0,0)[lt]{\lineheight{1.25}\smash{\begin{tabular}[t]{l}\textbf{bicycle}\end{tabular}}}}}%
    \put(0.95932299,0.14022433){\color[rgb]{0,0,0}\rotatebox{-90}{\makebox(0,0)[lt]{\lineheight{1.25}\smash{\begin{tabular}[t]{l}\textbf{agent$_{bicycle}$}\end{tabular}}}}}%
    \put(0.9834534,0.14022433){\color[rgb]{0,0,0}\rotatebox{-90}{\makebox(0,0)[lt]{\lineheight{1.25}\smash{\begin{tabular}[t]{l}\textbf{agent$_{motorcycle}$}\end{tabular}}}}}%
    \put(0.0360475,0.14383242){\color[rgb]{0,0,0}\makebox(0,0)[rt]{\lineheight{1.25}\smash{\begin{tabular}[t]{r}\textbf{1}\textbf{M}\end{tabular}}}}%
    \put(0.03604749,0.24603043){\color[rgb]{0,0,0}\makebox(0,0)[rt]{\lineheight{1.25}\smash{\begin{tabular}[t]{r}\textbf{10M}\end{tabular}}}}%
    \put(0.03604749,0.34822847){\color[rgb]{0,0,0}\makebox(0,0)[rt]{\lineheight{1.25}\smash{\begin{tabular}[t]{r}\textbf{100M}\end{tabular}}}}%
    \put(0.0360475,0.45042652){\color[rgb]{0,0,0}\makebox(0,0)[rt]{\lineheight{1.25}\smash{\begin{tabular}[t]{r}\textbf{1B}\end{tabular}}}}%
    \put(0.03604749,0.55262455){\color[rgb]{0,0,0}\makebox(0,0)[rt]{\lineheight{1.25}\smash{\begin{tabular}[t]{r}\textbf{10B}\end{tabular}}}}%
  \end{picture}%
\endgroup%

%% file: sec/4_eval.tex
\section{AnoVox Benchmark Suite}
\label{sec:eval}

We provide a full evaluation suite for camera and lidar-based anomaly detection methods for content anomalies. Based on pixel- or pointwise anomaly scores, we voxelize the results and compare them against the voxelized anomaly ground truth of AnoVox. We evaluate on voxel grid of size $100m \times 100m \times 64m$, where each voxel has a length of 0.5~m. We have performed ablation studies on smaller voxel sizes, but have not found a significant impact on the results.




\subsection{Anomaly Detection}
\label{subsec:detection}

To get first insights into how current state-of-the-art methods for \textit{content} anomalies will perform on our benchmark, we have implemented two candidate methods, one based on camera data and one based on lidar data. For their selection, we have analyzed existing surveys and benchmark results~\cite{Bogdoll_Anomaly_2022_CVPR,bogdoll_perception,Blum_Fishyscapes_2021_IJCV,Chan_SegmentMeIfYouCan_2021_NEURIPS}. However, as there is no existing benchmark for lidar-based detection method, this was more challenging. We performed an extensive search, also focusing on open-set segmentation, as this is a similar but more active field. Our final candidates can be found in Table~\ref{tab:sota_anodec}. To emphasize the definition of normality based on training data alone, we then neglected all methods that require auxiliary data with anomalies during training. While we are also interested in methods purely trained on raw data, we allowed the usage of semantic labels during training, as the state-of-the-art heavily relies on those. We then picked the best-performing method that provided open-source code for implementations. For camera data, we selected the Rejected By All~(RbA) method from Nayal et al.~\cite{nayal_rba_2023}, and for lidar data we selected the Redundancy classifier~(REAL) framework by Cen et al.~\cite{cen_open-world_2022}. While both methods allow for the usage of auxiliary data during training, we did not fine-tune on anomalies. For RbA this means we did not perform "Outlier Data Exposure"~\cite{nayal_rba_2023}, and for REAL this means we only used the "predictive distribution calibration" without "unknown object synthesis"~\cite{cen_open-world_2022}. 

\begin{figure}[t]
\begin{center}\resizebox{1\textwidth}{!}{\input{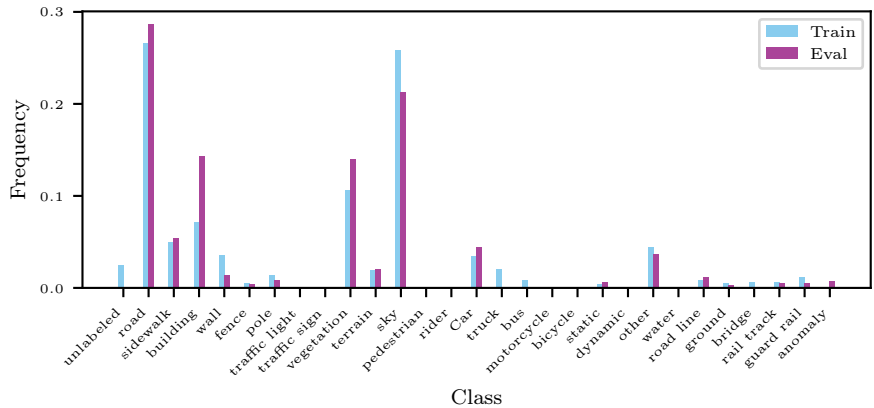}}\end{center}
\vspace*{-6mm}
\caption{Class distributions for the train and evaluation datasets.}
\label{fig:labels_train_eval}
\end{figure}

The \textbf{Rejected by All} model proposed by Nayal et al. uses camera data and is based on the Mask2Former model~\cite{cheng_masked-attention_2022}. The authors propose that object queries are specialized on single classes, such that outliers can be detected when the input is rejected by all queries. RbA was trained on Cityscapes and evaluated on SegmentMeIfYouCan anomaly and obstacle tracks, as well as on the Fishyscapes Lost\&Found track, where they perform well with FPR$_{95}$ values below 12. 

The \textbf{Redundancy classifier} approach from Cen et al. uses lidar data and is based on the Cylinder3D~\cite{zhu_cylindrical_2021} framework. To learn high anomaly scores, the authors introduce a new loss, where the second-highest predictions per point are assigned to an unknown class, which they use for uncertainty prediction during inference while also performing close-set semantic segmentation. RbA was trained and evaluated on both SemanticKITTI~\cite{behley_semantickitti_2019} and Nuscenes~\cite{caesar_nuscenes_2020}. The approach works reasonably well with an AUPR of 10.0 and an AUROC of 77.5. However, due to the domain gap, these values cannot easily be compared with camera-based methods.

\subsubsection{Training}
\label{subsec:detection_training}

For a fair comparison, we have trained both methods on the same training dataset. For this, we have created a small \textit{normality} instantiation in the size of the Cityscapes dataset that consists of 2,975 frames with temporal scenarios. Contrary to the setting of REAL, where unknown objects are included in the training set but ignored for the loss computation, our training dataset does not include any anomalies. We followed the standard training procedures and parameters as provided by the authors.


\begin{table}[t]
\centering
\caption{Evaluation of REAL and RbA on AnoVox.}
\label{tab:sota_eval}
\resizebox{0.6\textwidth}{!}{%
\begin{tabular}{lccccc}
\hline
                        & \textbf{AUPR} $\uparrow$        & \textbf{AUROC}$\uparrow$     & \textbf{F1}$\uparrow$ & \textbf{PPV}$\uparrow$ &
                        \textbf{FPR$_{95}$}$\downarrow$\\ \hline
REAL   & 0.14 & 43.30  & 0.0         & 0.0   & 100       \\
REAL$_{+norm}$          & 0.04                & 43.53            & 0.0         & 0.0   & 100       \\
REAL$_{big}$    & 0.17  & 44.7 & 0.0         & 0.0  & 100        \\
REAL$_{medium}$ & 0.11 & 42.8 & 0.0         & 0.0  & 100        \\
REAL$_{small}$  & 0.21 & 55.1   & 0.0         & 0.0 & 100         \\\midrule
RbA     & 0.7                   & 57.3               & 2.6         & 1.4    & 100      \\
RbA$_{+norm}$           & 0.2                   & 57.6               & 2.6         & 0.4  & 100        \\
RbA$_{big}$     & \textbf{2.3}                   & 54.9               & \textbf{3.7}         & \textbf{2.7}  & 100        \\
RbA$_{medium}$   & 0.6                   & \textbf{60.5}               & 0.0      & 0.0  & 100    \\
RbA$_{small}$    & 0.01                  & 53.2               & 0.0           & 0.0   & 100    \\\bottomrule 
\end{tabular}%
}
\end{table}

\subsubsection{Evaluation}
\label{subsec:detection_evaluation}

For the evaluation, we created another small instantiation with \textit{content} anomalies. The class distribution between the training and evaluation datasets can be found in Figure~\ref{fig:labels_train_eval}. To evaluate both methods, their anomaly scores need to be mapped to voxel space first. While this is straightforward for lidar data, we lifted the anomaly scores from RbA into 3D using ground truth depth data. Due to this evaluation in voxel space, it must be noted that the class imbalance between normal data and anomalies is much larger than in the sensor space due to quantization effects. Thus, results cannot be compared to reported values from the state of the Art. However, we have also performed evaluations on the sensor data directly, confirming higher scores. 

As shown in Table~\ref{tab:sota_eval}, both methods had significant issues detecting anomalies. While we observed stable training performance and saw improved performance when trained on larger datasets, the results were rather surprising. RbA was able to detect some anomalies, as shown in Fig.~\ref{fig:teaser}, but struggled with most. REAL on the other hand, while generating well-performing closed-set predictions, was unable to generate any meaningful uncertainties in the absence of unknown objects in the training data, as exemplarily shown in Fig.~\ref{fig:failures_ood}.

This shows that the task of anomaly detection becomes much harder on a challenging benchmark such as AnoVox, where anomalies are not defined by not belonging to Cityscapes classes, but are defined by being absent from the training set, as most state-of-the-art methods currently utilize deviations based on semantic segmentations to derive anomalies. In addition, there is no domain shift for the anomaly classes, as they are rendered in the same simulation engine from which the training data stems from. Finally, lots of frames without visible anomalies and particularly small anomalies raise the bar additionally. This setting, combined with the induced class imbalance due to our quantization loss during voxelization, makes it much harder to perform well on AnoVox compared to existing benchmarks, which we hope will benefit the community greatly.

%% file: sec/5_conclusion.tex
\section{Conclusion}
\label{sec:conclusion}

We presented the AnoVox benchmark for \textbf{ANO}maly detection in autonomous driving. AnoVox provides ground truth in the form of semantic masks for RGB images and lidar point clouds as well as spatial \textbf{VOX}els. 

\begin{figure}[t!]
\centering
    \resizebox{1\textwidth}{!}{
    \begin{subfigure}{0.33\textwidth}
        \includegraphics[width=\textwidth]{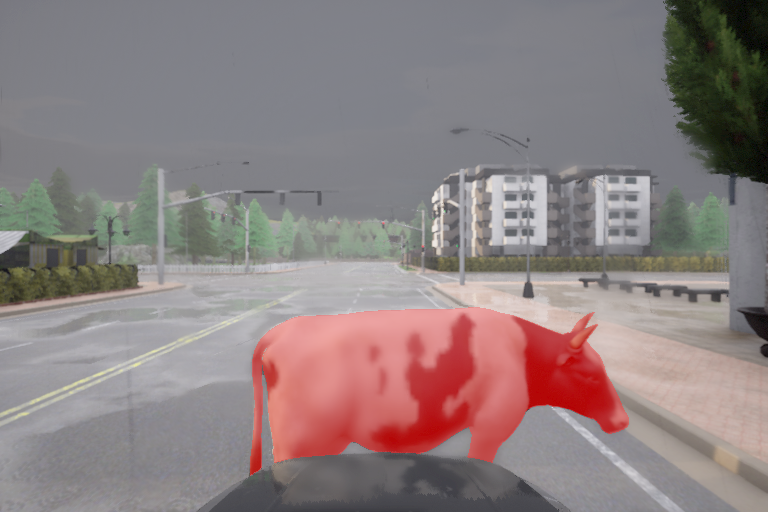}
    \end{subfigure}
    \begin{subfigure}{0.33\textwidth}
        \includegraphics[width=\textwidth]{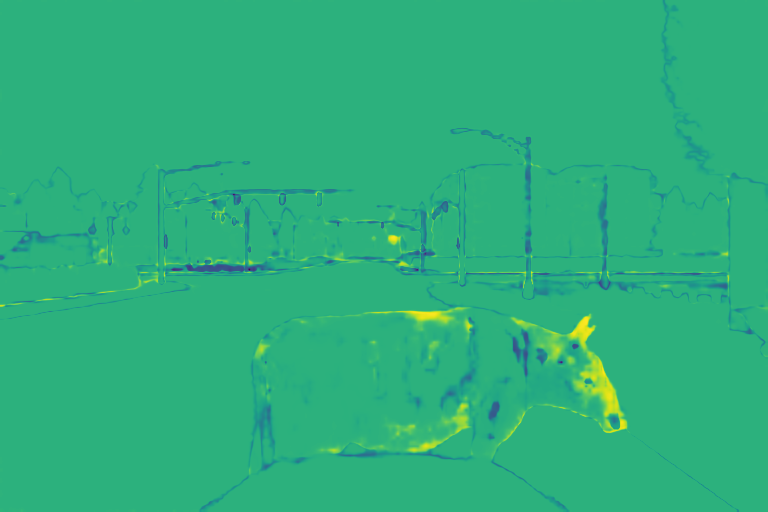}
    \end{subfigure}
    \begin{subfigure}{0.33\textwidth}
        \includegraphics[width=\textwidth]{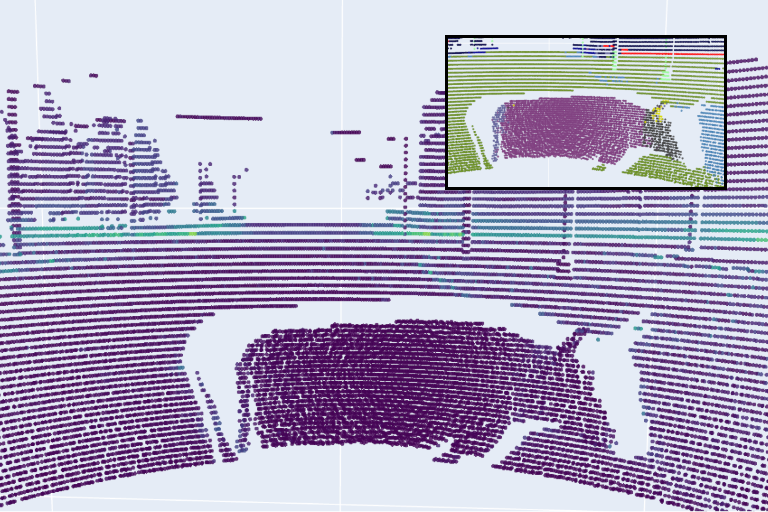}
    \end{subfigure}}
    \caption{AnoVox scene with a cow as an animal \textit{content} anomaly (left). The RbA (middle) anomaly detection method is only able to detect some border parts as anomalous, while REAL (right) assigns the same uncertainty to the cow as it does to the ground. In the accompanying closed-set detection, it is mostly classified as a car.}
  \label{fig:failures_ood}
\end{figure}

We analyzed the shortcomings of existing benchmarks, most notably no clear definition of normality and a strong focus on RGB data. AnoVox is the first and largest benchmark that provides a concise formalization of normality, accompanied by compliant training data. We provided first insights on how well current SotA models perform on this new paradigm of anomaly detection benchmarking. We see a great need for anomaly detection methods to learn normality based on training data instead of detecting deviations from known Cityscape classes.

\textbf{Limitations.} At the moment, each voxel contains only a single class. This can lead to cases where small anomalies are not represented in single frames of voxelized data. For our evaluation, we map camera-based anomaly detections into 3D by using ground truth depth. We have performed experiments using ZoeDepth~\cite{bhat_zoedepth_2023}, a state-of-the-art metric depth estimation model, to map these predictions into 3D, but the overlap between the resulting voxels and the ground truth voxels was too large. Here, more sophisticated methods are necessary. We limited ourselves to only a single temporal anomaly, more types could be implemented. Finally, while simulation is necessary to have full control over both training and evaluation, this leads to a  Sim-2-Real gap.


\textbf{Outlook.}
We plan to implement multi-class voxel in a future release to reduce losses during quantization. The Sim-2-Real gap could be tackled by style transfers, as proposed by Tian et al.~\cite{tian2024latencyaware}. We would like to include radar data in a future release, but the current implementation in CARLA is too naive to be helpful. Finally, we plan to release more data with our \textit{surround} sensor setup.

%% file: sec/6_acknowledgment.tex
\section*{Acknowledgment}
\label{sec:ackno}

This work results from the just better DATA project supported by the German Federal Ministry for Economic Affairs and Climate Action (BMWK), grant number 19A23003H.